\documentstyle[epsfig,times]{article}

\makeatletter
  \newlength\titlebox 
 \twocolumn 

\def\addcontentsline#1#2#3{}

\def\maketitle{\par
 \begingroup 
   \def\thefootnote{\fnsymbol{footnote}}
   \def\@makefnmark{\hbox to 0pt{$^{\@thefnmark}$\hss}}
   \twocolumn[\@maketitle] \@thanks
 \endgroup
\setcounter{footnote}{0}
 \let\maketitle\relax \let\@maketitle\relax
 \gdef\@thanks{}\gdef\@author{}\gdef\@title{}\let\thanks\relax}
\def\@maketitle{\vbox to \titlebox{\hsize\textwidth
 \linewidth\hsize \vskip 0.200in minus 0.125in \centering
 {\LARGE\bf \@title \par} \vskip 0.2in plus 1fil minus 0.1in
 {\def\and{\unskip\enspace{\rm and}\enspace}%
  \def\And{\end{tabular}\hss \egroup \hskip 1in plus 2fil 
           \hbox to 0pt\bgroup\hss \begin{tabular}[t]{c}\Large\bf}%
  \def\AND{\end{tabular}\hss\egroup \hfil\hfil\egroup
          \vskip 0.25in plus 1fil minus 0.125in
           \hbox to \linewidth\bgroup\Large \hfil\hfil
             \hbox to 0pt\bgroup\hss \begin{tabular}[t]{c}\Large\bf}
  \hbox to \linewidth\bgroup\normalsize \hfil\hfil
    \hbox to 0pt\bgroup\hss \begin{tabular}[t]{c}\Large\bf\@author 
                            \end{tabular}\hss\egroup
    \hfil\hfil\egroup}
  \vskip 0.3in plus 2fil minus 0.1in
}}

\renewenvironment{abstract}{\centerline{\bf
Abstract}\vspace{0.5ex}\begin{quote}\small}{\par\end{quote}\vskip 1ex}

\def\section{\@startsection {section}{1}{\z@}{-2.0ex plus
    -0.5ex minus -.2ex}{3pt plus 2pt minus 1pt}{\Large\bf\centering}}
\def\subsection{\@startsection{subsection}{2}{\z@}{-2.0ex plus
    -0.5ex minus -.2ex}{3pt plus 2pt minus 1pt}{\large\bf\raggedright}}
\def\subsubsection{\@startsection{subparagraph}{3}{\z@}{-6pt plus
   2pt minus 1pt}{-1em}{\normalsize\bf}}
\skip\footins 9pt plus 4pt minus 2pt
\def\footnoterule{\kern-3pt \hrule width 5pc \kern 2.6pt }
\labelwidth\leftmargini\advance\labelwidth-\labelsep \labelsep 5pt

\def\@listi{\leftmargin\leftmargini}
\def\@listii{\leftmargin\leftmarginii
   \labelwidth\leftmarginii\advance\labelwidth-\labelsep
   \topsep 2pt plus 1pt minus 0.5pt
   \parsep 1pt plus 0.5pt minus 0.5pt
   \itemsep \parsep}
\def\@listiii{\leftmargin\leftmarginiii
    \labelwidth\leftmarginiii\advance\labelwidth-\labelsep
    \topsep 1pt plus 0.5pt minus 0.5pt 
    \parsep \z@ \partopsep 0.5pt plus 0pt minus 0.5pt
    \itemsep \topsep}
\def\@listiv{\leftmargin\leftmarginiv
     \labelwidth\leftmarginiv\advance\labelwidth-\labelsep}
\def\@listv{\leftmargin\leftmarginv
     \labelwidth\leftmarginv\advance\labelwidth-\labelsep}
\def\@listvi{\leftmargin\leftmarginvi
     \labelwidth\leftmarginvi\advance\labelwidth-\labelsep}

\belowdisplayskip \abovedisplayskip
\def\@normalsize{\@setsize\normalsize{11pt}\xpt\@xpt}	
\def\small{\@setsize\small{10pt}\ixpt\@ixpt}		
\def\footnotesize{\@setsize\footnotesize{10pt}\ixpt\@ixpt} 
\def\scriptsize{\@setsize\scriptsize{8pt}\viipt\@viipt}	
\def\tiny{\@setsize\tiny{7pt}\vipt\@vipt}		
\def\large{\@setsize\large{12pt}\xipt\@xipt} 		
\def\Large{\@setsize\Large{14pt}\xiipt\@xiipt} 		
\def\LARGE{\@setsize\LARGE{16pt}\xivpt\@xivpt} 		
\def\huge{\@setsize\huge{20pt}\xviipt\@xviipt} 		
\def\Huge{\@setsize\Huge{23pt}\xxpt\@xxpt} 		

\def\leftcite{(}\def\rightcite{)}

\def\cite{\def\citeauthoryear##1##2{\def\@thisauthor{##1}%
 	     \ifx \@lastauthor \@thisauthor \relax \else##1 \fi ##2}\@icite}
\def\shortcite{\def\citeauthoryear##1##2{##2}\@icite}

\def\citeauthor{\def\citeauthoryear##1##2{##1}\@nbcite}
\def\citeyear{\def\citeauthoryear##1##2{##2}\@nbcite}

\def\@icite{\leavevmode\def\@citeseppen{-1000}%
 \def\@cite##1##2{\leftcite\nobreak\hskip 0in{##1\if@tempswa , ##2\fi}\rightcite}%
 \@ifnextchar [{\@tempswatrue\@citex}{\@tempswafalse\@citex[]}}
\def\@nbcite{\leavevmode\def\@citeseppen{1000}%
 \def\@cite##1##2{{##1\if@tempswa , ##2\fi}}%
 \@ifnextchar [{\@tempswatrue\@citex}{\@tempswafalse\@citex[]}}

\def\@citex[#1]#2{%
  \def\@lastauthor{}\def\@citea{}%
  \@cite{\@for\@citeb:=#2\do
    {\@citea\def\@citea{;\penalty\@citeseppen\ }%
     \if@filesw\immediate\write\@auxout{\string\citation{\@citeb}}\fi
     \@ifundefined{b@\@citeb}{\def\@thisauthor{}{\bf ?}\@warning
       {Citation `\@citeb' on page \thepage \space undefined}}%
     {\csname b@\@citeb\endcsname}\let\@lastauthor\@thisauthor}}{#1}}

\def\@lbibitem[#1]#2{\item\if@filesw 
      { \def\protect##1{\string ##1\space}\immediate
        \write\@auxout{\string\bibcite{#2}{#1}}}\fi\ignorespaces}

\def\thebibliography#1{\section*{References\@mkboth
 {REFERENCES}{REFERENCES}}\list
 {}{\labelwidth 0in\leftmargin\labelwidth
 \advance\leftmargin\labelsep \itemsep .01in}
 \def\newblock{\hskip .11em plus .33em minus .07em}
 \sloppy\clubpenalty4000\widowpenalty4000
 \sfcode`\.=1000\relax}

\makeatother

\newtheorem{example1}{Example}
\newenvironment{example}[1]{\begin{example1}[#1] \small\mbox{} \\ \rm}{\end{example1}}

\makeatletter
\def\ps@firstpage{
 \def\@oddfoot{\raisebox{-4em}{Presented at the AAAI Spring Symposium on Applying Machine Learning and Discourse Processing, Stanford, March 1998.}}
 \def\@oddhead{}
 \def\@evenfoot\@oddfoot
 \def\@evenhead{}
}
\makeatother

\begin{document}
\thispagestyle{firstpage}

\setlength\titlebox{1.6in}
\title{Identifying Discourse Markers in Spoken Dialog}

\author{Peter A. Heeman$^\dag$ \and Donna Byron$\ddag$ \and James F. Allen$^\ddag$ \vspace{0.5em}\\
\begin{tabular}{cc}
\makebox[22em][c]{$^\dag$Computer Science and Engineering} &
\makebox[22em][c]{$^\ddag$Department of Computer Science} \\
Oregon Graduate Institute &
University of Rochester \\
PO Box 91000 Portland OR 97291 &
Rochester NY 14627 \\
{\tt heeman@cse.ogi.edu} & 
{\tt \{dbyron,james\}@cs.rochester.edu}
\end{tabular}}

\maketitle

\begin{abstract}

In this paper, we present a method for identifying discourse marker
usage in spontaneous speech based on machine learning.  Discourse
markers are denoted by special POS tags, and thus the process of POS
tagging can be used to identify discourse markers.  By incorporating
POS tagging into language modeling, discourse markers can be
identified during speech recognition, in which the timeliness of the
information can be used to help predict the following words.  We
contrast this approach with an alternative machine learning approach
proposed by Litman \shortcite{Litman96:jair}.  This paper also argues
that discourse markers can be used to help the hearer predict the role
that the upcoming utterance plays in the dialog.  Thus discourse
markers should provide valuable evidence for automatic dialog act
prediction.

\end{abstract}

\newcommand{\comment}[1]{}
\newcommand{\pnote}[1]{\footnote{{\bf Peter:} #1}}

\newcommand{\mc}[1]{\multicolumn{1}{c|}{#1}}

\newcommand{\interruptionpoint}{
\parbox[t]{0.7em}{
\raisebox{-1.4em}{\hspace{-.4em}\LARGE\bf$\uparrow$} \vspace{-0.2em} \\
\makebox[1em][c]{\small\em interruption point} 
}}

\newcommand{\ip}{
\parbox[t]{0.5em}{
\raisebox{-1.4em}{\hspace{-.4em}\LARGE\bf$\uparrow$} \vspace{-0.4em} \\
\hspace*{-0.4em}{\small\em ip}}}

\newcommand{\reparandum}[1]{$\underbrace{\makebox{#1}}_{\makebox{\small\em reparandum}}$}
\newcommand{\editingterm}[1]{$\underbrace{\makebox{#1}}_{\makebox{\small\em
editing term}}$}
\newcommand{\et}[1]{$\underbrace{\makebox{#1}}_{\makebox{\small\em
et}}$}
\newcommand{\alteration}[1]{$\underbrace{\makebox{#1}}_{\makebox{\small\em alteration}}$}

\newcommand{\etal}{{\em et al.}}
\newcommand{\dash}{\mbox{-}}
\newcommand{\rim}{1\!,i\mbox{-}1}
\newcommand{\ri}{1\!,i}

\section{Introduction}

{\em Discourse markers} are a linguistic devise that speakers use to
signal how the upcoming unit of speech or text relates to the current
discourse state \cite {Schiffrin87:book}.  Previous work in
computational linguistics has emphasized their role in marking changes
in the global discourse structure (e.g.~\cite
{GroszSidner86:cl,Reichman85:book,RCohen84:coling}).  For instance,
``by the way'' is used to mark the start of a digression, ``anyway''
to mark the return from one, and ``now'' to shift to a new topic.
Schiffrin's work in social dialogue \shortcite{Schiffrin87:book} took
a much wider scope, and examined how discourse markers in general are
used.  She found that they are used to mark the information status in
an utterance and how it relates to the previous discourse state.  For
instance, when someone is about to disagree with information in the
discourse state, they might introduce the utterance with ``well''.

In human-human task-oriented dialogs, discourse markers abound.  In
the Trains corpus of spontaneous speech \cite{HeemanAllen95:cdrom},
44.1\% of the turns (other than acknowledgments) are introduced
with a discourse marker.  Because discourse markers are so prominent
in task-oriented dialogs, they could be a valuable source of
information for understanding the utterances that they introduce. This
striking feature of task-oriented dialog has been largely ignored by
other researchers in building spoken dialog systems, which simply
regard them as noise (cf.~\cite {DahlbackJonsson92:cogsci}).
Task-oriented dialogs manifest a considerably different surface form
than either monologues, social dialog or written text \cite
{BrownYule83:book}, so it is not clear whether discourse markers are
playing the same role in task-oriented dialogs as in other forms of
discourse.

One problem with discourse markers, however, is that there is
ambiguity as to whether lexical items are functioning as discourse
markers.  Consider the lexical item ``so''.  Not only can it be used
as a discourse marker to introduce an utterance, but it can also be
used sententially to indicate a subordinating clause as illustrated by
the following example from the Trains corpus.
\begin{example}{d93-15.2 utt9}
it takes an hour to load them \\
just so you know
\end{example}
Discourse markers can also be used inside an utterance to mark a {\em
speech repair}, where the speaker goes back and repeats or corrects
something she just said.  Here, the discourse markers play a much more
internal role, as the following example with ``well'' illustrates.
\begin{example}{d93-26.3 utt12}
\reparandum{can I have engine}\ip \et{well} if I take engine one and pick up a boxcar 
\end{example}
Due to these difficulties, an effective algorithm for identifying
discourse markers in spontaneous speech needs to also address the
problem of segmenting speech into utterance units and identifying
speech repairs \cite {HeemanAllen97:acl}.

In the rest of this paper, we first review the Trains corpus and the
manner in which the discourse markers were annotated by using special
part-of-speech (POS) tags to denote them.  We then examine the role
that discourse markers play in task-oriented dialogs.  We then present
our speech recognition language model, which incorporates POS tagging,
and thus discourse marker identification.  We show that distinguishing
discourse marker usages results in improved language modeling.  We
also show that discourse marker identification is improved by modeling
interactions with utterance segmentation and resolving speech repairs.
From this, we conclude that discourse markers can be used by hearers
to set up expectations of the role that the upcoming utterance plays
in the dialog.  Due to the ability to automatically identify discourse
markers during the speech recognition process, we argue that they can
be exploited in the task of dialog act identification, which is
currently receiving much attention in spontaneous speech research
(e.g.~\cite
{Taylor-etal97:eurospeech,Chucarroll98:aaaiss,Stolcke-etal98:aaaiss}).
We conclude with a comparison to the method proposed by Litman
\shortcite {Litman96:jair} for identifying discourse markers.

\section{Trains Corpus}
\label{sec:corpus}

As part of the Trains project \cite{Allen-etal95:jetai-s}, which is a
long term research project to build a conversationally proficient
planning assistant, we have collected a corpus of problem solving
dialogs \cite{HeemanAllen95:cdrom}.  The dialogs involve two human
participants, one who is playing the role of a user and has a certain
task to accomplish, and another who is playing the role of the system
by acting as a planning assistant.  The collection methodology was
designed to make the setting as close to human-computer interaction as
possible, but was not a {\em wizard} scenario, where one person
pretends to be a computer; rather, the user knows that he is talking
to another person.  The Trains corpus consists of approximately six
and a half hours of speech. Table~\ref{tab:occurrences} gives some
general statistics about the corpus, including the number of dialogs,
speakers, words, speaker turns, and occurrences of discourse markers.
\begin{table}[h]
\begin{center}
{\small\begin{tabular}{|l|r|}\hline
Dialogs                      &    98 \\
Speakers                     &    34 \\
Words                        & 58298 \\
Turns                        &  6163 \\
Discourse Markers            &  8278 \\ \hline
\end{tabular}}
\end{center}
\caption{Size of the Trains Corpus}
\label{tab:occurrences}
\end{table}

Our strategy for annotating discourse markers is to mark such usages
with special POS tags.  Four special POS tags were added to the Penn
Treebank tagset \cite {Marcus-etal93:cl} to denote discourse marker
usage.  These tags are defined in Table~\ref{tab:dm_defs}.\footnote
{Other additions to the tagset are described in Heeman
\shortcite{Heeman97:thesis}.}
\begin{table}[h]
\begin{description}\setlength\itemsep{-.1cm}
\item[AC:] Single word acknowledgments, such as ``okay'', ``right'', ``mm-hm'',
``yeah'', ``yes'', ``alright'', ``no'', and ``yep''. 
\item[UH\_D:] Interjections with discourse purpose, such as ``oh'''', ``well'', ``hm'', ``mm'', and ``like''.
\item[CC\_D:] Co-ordinating conjuncts used as discourse markers, such as
``and'', ``so'', ``but'', ``oh'', and ``because''.
\item[RB\_D:] Adverbials used as discourse markers, such as ``then'', ``now'',
``actually'', ``first'', and ``anyway''.
\end{description}\vspace*{-1em}
\caption{\label{tab:dm_defs}POS tags for Discourse Markers}
\end{table}
Verbs used as discourse markers, such as ``wait'', and ``see'', are
not given special markers, but are annotated as verbs.  Also, no
attempt has been made at analyzing multi-word discourse markers, such
as ``by the way'' and ``you know''.  However, phrases such as ``oh
really'' and ``and then'' are treated as two individual discourse
markers.  Lastly, filled pause words, namely ``uh'', ``um'' and
``er'', are marked with {\bf UH\_FP}; but these are not considered as
discourse markers.

\section{POS-Based Language Model}

The traditional goal of speech recognition is to find the sequence of
words $\hat{W}$ that is maximal given the acoustic signal $A$.  In
earlier work \cite{HeemanAllen97:eurospeech,Heeman97:thesis}, we argue
that this view is too limiting.  In a spoken dialog system, word
recognition is just the first step in understanding the speaker's
turn.  Furthermore, speech recognition is difficult especially
without the use of higher level information.  Hence, we propose as a
first step to incorporate POS tagging into the speech recognition
process.

Previous approaches that have made use of POS tags in speech
recognition view the POS tags as intermediate objects by summing over
the POS tag sequences \cite {Jelinek85}.  Instead, we take the
approach of redefining the goal of the speech recognition process so
that it finds the best word ($W$) and POS tag ($P$) sequence given the
acoustic signal.  The derivation of the acoustic model and language
model is now as follows.
\begin{eqnarray*}
\hat{W}\hat{P}&=&\arg\max_{W,P}\Pr(WP|A)\\
&=&\arg\max_{WP}\frac{\Pr(A|WP)\Pr(WP)}{\Pr(A)}\\ 
&=&\arg\max_{WP}\Pr(A|WP)\Pr(WP)
\end{eqnarray*}
The first term $\Pr(A|WP)$ is the factor due to the acoustic model,
which we can approximate by $\Pr(A|W)$. The second term $\Pr(WP)$ is
the factor due to the language model. We rewrite $\Pr(WP)$ as
$\Pr(W_{1,N}P_{1,N})$, where $N$ is the number of words in the
sequence.  We now rewrite the language model probability as follows.
\begin{eqnarray*}
\lefteqn{\Pr(W_{1,N}P_{1,N})} \\
&=&\prod_{i=1,N}\Pr(W_iP_i|W_{\rim}P_{\rim})\\
&=&\prod_{i=1,N}\Pr(W_i|W_{\rim}P_{\ri})\Pr(P_i|W_{\rim}P_{\rim})
\end{eqnarray*}
The final probability distributions are similar to those used by
previous attempts to use POS tags in language modeling \cite
{Jelinek85} and those used for POS tagging of written text \cite
{Charniak-etal93:aaai,Church88:anlp,DeRose88:cl}.  However, these
approaches simplify the probability distributions as shown by the
approximations below.
\begin{eqnarray*}
\Pr(W_i|W_{\rim}P_{\ri}) &\approx& \Pr(W_i|P_i) \\
\Pr(P_i|W_{\rim}P_{\rim}) &\approx& \Pr(P_i|P_{\rim})
\end{eqnarray*}
However, as we have shown in earlier work \cite
{HeemanAllen97:eurospeech,Heeman97:thesis}, such simplifications lead
to poor language models.

\subsection{Probability Distributions}

We have two probability distributions that need to be estimated.  The
simplest approach for estimating the probability of an event given a
context is to use the relative frequency that the event occurs given
the context according to a training corpus.  However, no matter how
large the training corpus is, there will always be event-context pairs
that have not been seen or that have been seen too rarely to
accurately estimate the probability.  To alleviate this problem, one
can partition the contexts into a smaller number of equivalence
classes and use these equivalence classes to compute the relative
frequencies.

We use a decision tree learning algorithm \cite
{Bahl-etal89:tassp,Black-etal92:darpa:pos,Breiman-etal84:book}, which
uses information theoretic measures to construct equivalence classes
of the context in order to cope with sparseness of data. The decision
tree algorithm starts with all of the training data in a single leaf
node. For each leaf node, it looks for the question to ask of the
context such that splitting the node into two leaf nodes results in
the biggest decrease in {\em impurity}, where the impurity measures
how well each leaf predicts the events in the node.  Heldout data is
used to decide when to stop growing the tree: a split is rejected if
the split does not result in a decrease in impurity with respect to
the heldout data. After the tree is grown, the heldout dataset is used
to smooth the probabilities of each node with its parent \cite
{Bahl-etal89:tassp}.

\begin{figure}
\begin{center}
\begin{picture}(0,0)%
\includegraphics{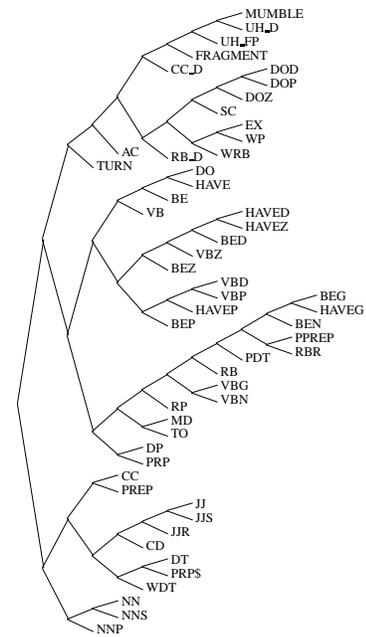}%
\end{picture}%
\setlength{\unitlength}{0.008125in}%
\begingroup\makeatletter\ifx\SetFigFont\undefined
\def\x#1#2#3#4#5#6#7\relax{\def\x{#1#2#3#4#5#6}}%
\expandafter\x\fmtname xxxxxx\relax \def\y{splain}%
\ifx\x\y   
\gdef\SetFigFont#1#2#3{%
  \ifnum #1<17\tiny\else \ifnum #1<20\small\else
  \ifnum #1<24\normalsize\else \ifnum #1<29\large\else
  \ifnum #1<34\Large\else \ifnum #1<41\LARGE\else
     \huge\fi\fi\fi\fi\fi\fi
  \csname #3\endcsname}%
\else
\gdef\SetFigFont#1#2#3{\begingroup
  \count@#1\relax \ifnum 25<\count@\count@25\fi
  \def\x{\endgroup\@setsize\SetFigFont{#2pt}}%
  \expandafter\x
    \csname \romannumeral\the\count@ pt\expandafter\endcsname
    \csname @\romannumeral\the\count@ pt\endcsname
  \csname #3\endcsname}%
\fi
\fi\endgroup
\begin{picture}(193,413)(1,422)
\put(146,819){\makebox(0,0)[lb]{\smash{\SetFigFont{5}{6.0}{rm} MUMBLE}}}
\put(146,810){\makebox(0,0)[lb]{\smash{\SetFigFont{5}{6.0}{rm} UH\_D}}}
\put(130,801){\makebox(0,0)[lb]{\smash{\SetFigFont{5}{6.0}{rm} UH\_FP}}}
\put(114,792){\makebox(0,0)[lb]{\smash{\SetFigFont{5}{6.0}{rm} FRAGMENT}}}
\put( 98,783){\makebox(0,0)[lb]{\smash{\SetFigFont{5}{6.0}{rm} CC\_D}}}
\put(162,783){\makebox(0,0)[lb]{\smash{\SetFigFont{5}{6.0}{rm} DOD}}}
\put(162,774){\makebox(0,0)[lb]{\smash{\SetFigFont{5}{6.0}{rm} DOP}}}
\put(146,765){\makebox(0,0)[lb]{\smash{\SetFigFont{5}{6.0}{rm} DOZ}}}
\put(130,756){\makebox(0,0)[lb]{\smash{\SetFigFont{5}{6.0}{rm} SC}}}
\put(146,747){\makebox(0,0)[lb]{\smash{\SetFigFont{5}{6.0}{rm} EX}}}
\put(146,738){\makebox(0,0)[lb]{\smash{\SetFigFont{5}{6.0}{rm} WP}}}
\put(130,729){\makebox(0,0)[lb]{\smash{\SetFigFont{5}{6.0}{rm} WRB}}}
\put( 98,727){\makebox(0,0)[lb]{\smash{\SetFigFont{5}{6.0}{rm} RB\_D}}}
\put( 66,730){\makebox(0,0)[lb]{\smash{\SetFigFont{5}{6.0}{rm} AC}}}
\put( 50,721){\makebox(0,0)[lb]{\smash{\SetFigFont{5}{6.0}{rm} TURN}}}
\put(114,718){\makebox(0,0)[lb]{\smash{\SetFigFont{5}{6.0}{rm} DO}}}
\put(114,709){\makebox(0,0)[lb]{\smash{\SetFigFont{5}{6.0}{rm} HAVE}}}
\put( 98,700){\makebox(0,0)[lb]{\smash{\SetFigFont{5}{6.0}{rm} BE}}}
\put( 82,691){\makebox(0,0)[lb]{\smash{\SetFigFont{5}{6.0}{rm} VB}}}
\put(146,691){\makebox(0,0)[lb]{\smash{\SetFigFont{5}{6.0}{rm} HAVED}}}
\put(146,682){\makebox(0,0)[lb]{\smash{\SetFigFont{5}{6.0}{rm} HAVEZ}}}
\put(130,673){\makebox(0,0)[lb]{\smash{\SetFigFont{5}{6.0}{rm} BED}}}
\put(114,664){\makebox(0,0)[lb]{\smash{\SetFigFont{5}{6.0}{rm} VBZ}}}
\put( 98,655){\makebox(0,0)[lb]{\smash{\SetFigFont{5}{6.0}{rm} BEZ}}}
\put(130,646){\makebox(0,0)[lb]{\smash{\SetFigFont{5}{6.0}{rm} VBD}}}
\put(130,637){\makebox(0,0)[lb]{\smash{\SetFigFont{5}{6.0}{rm} VBP}}}
\put(114,628){\makebox(0,0)[lb]{\smash{\SetFigFont{5}{6.0}{rm} HAVEP}}}
\put( 98,619){\makebox(0,0)[lb]{\smash{\SetFigFont{5}{6.0}{rm} BEP}}}
\put(194,637){\makebox(0,0)[lb]{\smash{\SetFigFont{5}{6.0}{rm} BEG}}}
\put(194,628){\makebox(0,0)[lb]{\smash{\SetFigFont{5}{6.0}{rm} HAVEG}}}
\put(178,619){\makebox(0,0)[lb]{\smash{\SetFigFont{5}{6.0}{rm} BEN}}}
\put(178,610){\makebox(0,0)[lb]{\smash{\SetFigFont{5}{6.0}{rm} PPREP}}}
\put(178,601){\makebox(0,0)[lb]{\smash{\SetFigFont{5}{6.0}{rm} RBR}}}
\put(146,597){\makebox(0,0)[lb]{\smash{\SetFigFont{5}{6.0}{rm} PDT}}}
\put(130,588){\makebox(0,0)[lb]{\smash{\SetFigFont{5}{6.0}{rm} RB}}}
\put(130,579){\makebox(0,0)[lb]{\smash{\SetFigFont{5}{6.0}{rm} VBG}}}
\put(130,570){\makebox(0,0)[lb]{\smash{\SetFigFont{5}{6.0}{rm} VBN}}}
\put( 98,566){\makebox(0,0)[lb]{\smash{\SetFigFont{5}{6.0}{rm} RP}}}
\put( 98,557){\makebox(0,0)[lb]{\smash{\SetFigFont{5}{6.0}{rm} MD}}}
\put( 98,548){\makebox(0,0)[lb]{\smash{\SetFigFont{5}{6.0}{rm} TO}}}
\put( 82,539){\makebox(0,0)[lb]{\smash{\SetFigFont{5}{6.0}{rm} DP}}}
\put( 82,530){\makebox(0,0)[lb]{\smash{\SetFigFont{5}{6.0}{rm} PRP}}}
\put( 66,521){\makebox(0,0)[lb]{\smash{\SetFigFont{5}{6.0}{rm} CC}}}
\put( 66,512){\makebox(0,0)[lb]{\smash{\SetFigFont{5}{6.0}{rm} PREP}}}
\put(114,503){\makebox(0,0)[lb]{\smash{\SetFigFont{5}{6.0}{rm} JJ}}}
\put(114,494){\makebox(0,0)[lb]{\smash{\SetFigFont{5}{6.0}{rm} JJS}}}
\put( 98,485){\makebox(0,0)[lb]{\smash{\SetFigFont{5}{6.0}{rm} JJR}}}
\put( 82,476){\makebox(0,0)[lb]{\smash{\SetFigFont{5}{6.0}{rm} CD}}}
\put( 98,467){\makebox(0,0)[lb]{\smash{\SetFigFont{5}{6.0}{rm} DT}}}
\put( 98,458){\makebox(0,0)[lb]{\smash{\SetFigFont{5}{6.0}{rm} PRP\$}}}
\put( 82,449){\makebox(0,0)[lb]{\smash{\SetFigFont{5}{6.0}{rm} WDT}}}
\put( 66,440){\makebox(0,0)[lb]{\smash{\SetFigFont{5}{6.0}{rm} NN}}}
\put( 66,431){\makebox(0,0)[lb]{\smash{\SetFigFont{5}{6.0}{rm} NNS}}}
\put( 50,422){\makebox(0,0)[lb]{\smash{\SetFigFont{5}{6.0}{rm} NNP}}}
\end{picture}
\end{center}
\caption{POS Classification Tree}
\label{fig:postags}
\end{figure}

To allow the decision tree to ask questions about the words and POS
tags in the context such that the questions can generalize about words
and POS tags that behave similarly, we cluster the words and POS tags
using the algorithm of Brown~\etal~\shortcite {Brown-etal92:cl} into a
binary classification tree.  The algorithm starts with each word (or
POS tag) in a separate class, and successively merges classes that
result in the smallest lost in mutual information in terms of the
co-occurrences of these classes.  By keeping track of the order that
classes were merged, we can construct a hierarchical classification of
the classes.  Figure~\ref{fig:postags} shows a POS classification
tree, which was automatically built from the training data.
%
Note that the classification algorithm has clustered the discourse
marker POS tags close to each other in the classification tree.

The binary classification tree gives an implicit binary encoding for
each POS tag, which is determined by the sequence of top and bottom
edges that leads from the root node to the node for the POS tag.  The
binary encoding allows the decision tree to ask about the words and
POS tags using simple binary questions, such as `is the third bit of
the POS tag encoding equal to one?'  the POS tag.

\begin{figure}
\hspace{2em}
\begin{picture}(0,0)%
\epsfig{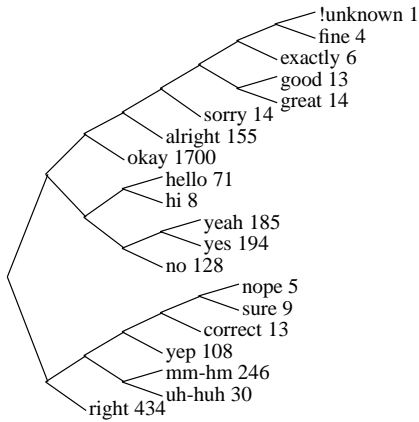}%
\end{picture}%
\setlength{\unitlength}{0.01250000in}%
\begingroup\makeatletter\ifx\SetFigFont\undefined
\def\x#1#2#3#4#5#6#7\relax{\def\x{#1#2#3#4#5#6}}%
\expandafter\x\fmtname xxxxxx\relax \def\y{splain}%
\ifx\x\y   
\gdef\SetFigFont#1#2#3{%
  \ifnum #1<17\tiny\else \ifnum #1<20\small\else
  \ifnum #1<24\normalsize\else \ifnum #1<29\large\else
  \ifnum #1<34\Large\else \ifnum #1<41\LARGE\else
     \huge\fi\fi\fi\fi\fi\fi
  \csname #3\endcsname}%
\else
\gdef\SetFigFont#1#2#3{\begingroup
  \count@#1\relax \ifnum 25<\count@\count@25\fi
  \def\x{\endgroup\@setsize\SetFigFont{#2pt}}%
  \expandafter\x
    \csname \romannumeral\the\count@ pt\expandafter\endcsname
    \csname @\romannumeral\the\count@ pt\endcsname
  \csname #3\endcsname}%
\fi
\fi\endgroup
\begin{picture}(133,181)(23,652)
\put(154,820){\makebox(0,0)[lb]{\smash{\SetFigFont{8}{9.6}{rm} !unknown 1}}}
\put(154,811){\makebox(0,0)[lb]{\smash{\SetFigFont{8}{9.6}{rm} fine 4}}}
\put(138,802){\makebox(0,0)[lb]{\smash{\SetFigFont{8}{9.6}{rm} exactly 6}}}
\put(138,793){\makebox(0,0)[lb]{\smash{\SetFigFont{8}{9.6}{rm} good 13}}}
\put(138,784){\makebox(0,0)[lb]{\smash{\SetFigFont{8}{9.6}{rm} great 14}}}
\put(106,778){\makebox(0,0)[lb]{\smash{\SetFigFont{8}{9.6}{rm} sorry 14}}}
\put( 90,769){\makebox(0,0)[lb]{\smash{\SetFigFont{8}{9.6}{rm} alright 155}}}
\put( 74,760){\makebox(0,0)[lb]{\smash{\SetFigFont{8}{9.6}{rm} okay 1700}}}
\put( 90,751){\makebox(0,0)[lb]{\smash{\SetFigFont{8}{9.6}{rm} hello 71}}}
\put( 90,742){\makebox(0,0)[lb]{\smash{\SetFigFont{8}{9.6}{rm} hi 8}}}
\put(106,733){\makebox(0,0)[lb]{\smash{\SetFigFont{8}{9.6}{rm} yeah 185}}}
\put(106,724){\makebox(0,0)[lb]{\smash{\SetFigFont{8}{9.6}{rm} yes 194}}}
\put( 90,715){\makebox(0,0)[lb]{\smash{\SetFigFont{8}{9.6}{rm} no 128}}}
\put(122,706){\makebox(0,0)[lb]{\smash{\SetFigFont{8}{9.6}{rm} nope 5}}}
\put(122,697){\makebox(0,0)[lb]{\smash{\SetFigFont{8}{9.6}{rm} sure 9}}}
\put(106,688){\makebox(0,0)[lb]{\smash{\SetFigFont{8}{9.6}{rm} correct 13}}}
\put( 90,679){\makebox(0,0)[lb]{\smash{\SetFigFont{8}{9.6}{rm} yep 108}}}
\put( 90,670){\makebox(0,0)[lb]{\smash{\SetFigFont{8}{9.6}{rm} mm-hm 246}}}
\put( 90,661){\makebox(0,0)[lb]{\smash{\SetFigFont{8}{9.6}{rm} uh-huh 30}}}
\put( 58,655){\makebox(0,0)[lb]{\smash{\SetFigFont{8}{9.6}{rm} right 434}}}
\end{picture}
\caption{AC Classification Tree}
\label{fig:ac}
\end{figure}

Unlike other work (e.g.~\cite{Black-etal92:darpa:pos,Magerman95:acl}),
we treat the word identities as a further refinement of the POS tags;
thus we build a word classification tree for each POS tag.  We grow
the classification tree by starting with a unique class for each word
and each POS tag that it takes on. When we merge classes to form the
hierarchy, we only allow merges if all of the words in both classes
have the same POS tag. The result is a word classification tree for
each POS tag.  This approach of building a word classification tree
for each POS tag has the advantage that it better deals with words
that can take on multiple senses, such as the word ``loads'', which
can be a plural noun ({\bf NNS}) or a present tense third-person verb
({\bf VBZ}).  As well, it constrains the task of building the word
classification trees since the major distinctions are captured by the
POS classification tree, thus allowing us to build classification
trees even for small corpora.  Figure~\ref {fig:ac} gives the
classification tree for the acknowledgments ({\bf AC}).
%
%
For each word, we give the number of times that it occurred in the
training data.  Words that only occurred once in the training corpus
have been grouped together in the class `{\bf !unknown}'.  Although the
clustering algorithm was able to group some of the similar
acknowledgments with each other, such as the group of ``mm-hm'' and
``uh-huh'', the group of ``good'', ``great'', and ``fine'', other
similar words were not grouped together, such as ``yep'' with ``yes''
and ``yeah'', and ``no'' with ``nope''.  Word adjacency information is
insufficient for capturing such semantic information.

\section{Results}
\label{sec:results}

To demonstrate our model, we use a 6-fold cross validation procedure,
in which we use each sixth of the corpus for testing data, and the
rest for training data.  We start with the word transcriptions of the
Trains corpus, thus allowing us to get a clearer indication of the
performance of our model without having to take into account the poor
performance of speech recognizers on spontaneous speech. 

Table~\ref{tab:simple:dm} reports the results of explicitly modeling
discourse markers with special POS tags.  
\begin{table}[b]
\begin{center}\small
\setlength{\tabcolsep}{0.3em}
\begin{tabular}{|l|r|r|r|} \hline
                  & No DM & DM   \\ \hline \hline
POS Errors        &  1219 &  1189 \\
POS Error Rate    &  2.09 &  2.04 \\ 
Perplexity        & 24.20 & 24.04 \\ \hline
\end{tabular}
\end{center}
\caption{Discourse Markers and Perplexity}
\label{tab:simple:dm}
\end{table}%
The second column, ``No DM'', reports the results of collapsing the
discourse marker usages with the sentential usages.  Thus, the
discourse conjunct {\bf CC\_D} is collapsed into {\bf CC}, the
discourse adverbial {\bf RB\_D} is collapsed into {\bf RB}, and the
acknowledgment {\bf AC} and discourse interjection {\bf UH\_D} are
collapsed into {\bf UH\_FP}.  The third column gives the results of
the model that does distinguish discourse marker usages, but ignoring
POS errors due to miscategorizing words as being discourse markers or
not.  We see that modeling discourse markers results in a reduction of
POS errors from 1219 to 1189, giving a POS error rate of 2.04\%.  We
also see a small decrease in perplexity from 24.20 to 24.04.
Perplexity of a test set of $N$ words $w_{1,N}$ is calculated as
follows.
\[ 2^{-\frac1N\sum_{i=1}^N\log_2\Pr(w_i|w_{\rim})} \]

In previous work \cite{HeemanAllen97:acl,Heeman97:thesis}, we argued
that discourse marker identification is tightly intertwined with the
problems of intonational phrase identification and resolving speech
repairs.  These three tasks, we claim, are necessary in order to
understand the user's contributions.  In Table~\ref{tab:full:dm}, we
show how discourse marker identification, POS tagging and perplexity
benefit by modeling the speaker's utterance.  The second column gives
the results of the POS-based model, which was used in the third column
of Table~\ref {tab:simple:dm}, the third column gives the results of
incorporating the detection and correction of speech repairs and
detection of intonational phrase boundary tones, and the fourth column
gives the results of adding in silence information to give further
evidence as to whether a speech repair or boundary tone occurred.
\begin{table}
\setlength{\tabcolsep}{0.2em}
\begin{center}\small
{\small\begin{tabular}{|l|r|r|r|r|} \hline
    &           &                 &\mc{Tones}      \vspace{-0.2em}\\
    &           &\mc{Tones}       &\mc{Repairs}    \vspace{-0.2em}\\
    &\mc{Base}  &\mc{Repairs}     &\mc{Corrections}\vspace{-0.2em}\\
    &\mc{Model} &\mc{Corrections} &\mc{Silences}   \\ \hline \hline
{\em POS Tagging}      &       &       &       \\
\ Errors               &  1711 &  1652 &  1572 \\
\ Error Rate           &  2.93 &  2.83 &  2.69 \\ \hline
Perplexity             & 24.04 & 22.96 & 22.35 \\ \hline
{\em Discourse Markers}&       &       &       \\
\ Errors               &   630 &   611 &   533 \\
\ Recall               & 96.75 & 96.67 & 97.26 \\
\ Precision            & 95.68 & 95.97 & 96.32 \\ \hline
\end{tabular}}
\end{center}
\caption{POS Tagging and Perplexity Results}
\label{tab:full:dm}
\end{table}%
As can be seen, modeling the user's utterances improves POS tagging
and word perplexity; adding in silence information to help detect
speech repairs and intonational boundaries further improves these two
rates.\footnote{Note the POS results include errors due to
miscategorizing discourse markers, which were excluded from the POS
results reported in Table~\ref{tab:simple:dm}.}  Of concern to this
paper, we also see an improvement in the identification of discourse
markers, improving from 630 to 533 errors.  This gives a final recall
rate of 97.26\% and a precision of 96.32.\footnote{The recall rate is
the number of discourse markers that were correctly identified over
the actual number of discourse markers. The precision rate is the
number of correctly identified discourse markers over the total number
of discourse markers guessed.}  In Heeman \shortcite{Heeman97:thesis},
we also show that modeling discourse markers improves the detection of
speech repairs and intonational boundaries.

\section{Comparison to Other Work}

Hirschberg and Litman \shortcite {HirschbergLitman93:cl} examined how
intonational information can distinguish between the discourse and
sentential interpretation for a set of ambiguous lexical items.  This
work was based on hand-transcribed intonational features and examined
discourse markers that were one word long.  In an initial study of the
discourse marker ``now'', they found that discourse usages of the word
``now'' were either an intermediate phrase by themselves (or in a
phrase consisting entirely of ambiguous tokens), or they are first in
an intermediate phrase (or preceded by other ambiguous tokens) and are
either de-accented or have a low accent (${\bf L}^{\bf *}$).
Sentential uses were either non-initial in a phrase or, if first, bore
a high (${\bf H}^{\bf *}$) or complex accent (i.e.~not a ${\bf L}^{\bf *}$
accent).  In a second study, Hirschberg and Litman used a speech
consisting of approximately 12,500 words.  They found that the
intonational model that they had proposed for the discourse marker
``now'' achieved a recall rate of 63.1\% of the discourse markers with
a precision of 88.3\%.\footnote{See Heeman \shortcite{Heeman97:thesis}
for a derivation of the recall and precision rates.}

Hirschberg and Litman also looked at the effect of orthographic
markers and POS tags.  For the orthographic markings, they looked at
how well discourse markers can be predicted based on whether they
follow or precede a hand-annotated punctuation mark.  They also
examined correlations with POS tags.  For this experiment, rather than
define special POS tags as we have done, they choose discourse marker
interpretation versus sentential interpretation based on whichever is
more likely for that POS tag, where the POS tags were automatically
computed using Church's part-of-speech tagger \shortcite
{Church88:anlp}.  This gives them a recall rate of 39.0\% and a
precision of 55.2\%.

Litman \shortcite{Litman96:jair} explored using machine learning
techniques to automatically learn classification rules for discourse
markers.  She contrasted the performance of CGRENDEL \cite
{WCohen92:ml,WCohen93:ijcai} with C4.5 \cite {Quinlan93:book}.
CGRENDEL is a learning algorithm that learns an ordered set of if-then
rules that map a condition to its most-likely event (in this case
discourse or sentential interpretation of potential discourse marker).
C4.5 is a decision tree growing algorithm that learns a hierarchical
set of if-then rules in which the leaf nodes specify the mapping to
the most-likely event.  She found that machine learning techniques
could be used to learn a classification algorithm that was as good as
the algorithm manually built by Hirschberg and Litman \shortcite
{HirschbergLitman93:cl}.  Further improvements were obtained when
different sets of features about the context were explored, such as
the identity of the token under consideration.  The best results
(although the differences between this version and some of the others
might not be significant) were obtained by using CGRENDEL and letting
it choose conditions from the following set: length of intonational
phrase, position of token in intonational phrase, length of
intermediate phrase, position of token in intermediate phrase,
composition of intermediate phrase (token is alone in intermediate
phrase, phrase consists entirely of potential discourse markers, or
otherwise), and identity of potential discourse marker.  The
automatically derived classification algorithm achieved a success rate
of 85.5\%, which translates into a discourse marker error rate of
37.3\%, in comparison to the error rate of 45.3\% for the algorithm of
Hirschberg and Litman \shortcite {HirschbergLitman93:cl}.  Hence,
machine learning techniques are an effective way in which a number of
different sources of information can be combined to identify discourse
markers.

Direct comparisons with our results are problematic since our corpus
is approximately five times as large.  Also we use task-oriented
human-human dialogs, rather than a monologue, and hence our corpus
includes a lot of turn-initial discourse markers for co-ordinating
mutual belief.  However, our results are based on automatically
identifying intonational boundaries, rather than including these as
part of the input.  In any event, the work of Litman and the earlier
work with Hirschberg indicate that our results can be further improved
by also modeling intermediate phrase boundaries (phrase accents), and
word accents, and by improving our modeling of these events, perhaps
by using more acoustic cues.  Conversely, we feel
that our approach, which integrates discourse marker identification
with speech recognition along with POS tagging, boundary tone
identification and the resolution of speech repairs, allows different
interpretations to be explored in parallel, rather than forcing
individual decisions to be made about each ambiguous token.  This
allows interactions between these problems to be modeled, which we
feel accounts for some of the improvement between our results and the
results reported by Litman.

\section{Predicting Speech Acts}
\label{sec:prediction}

Discourse markers are a prominent feature of human-human task-oriented
dialogs.  In this section, we examine the role that discourse markers,
other than acknowledgments, play at the beginning of speaker turns and
show that discourse markers can be used by the hearer to set up
expectations of the role that the upcoming utterance plays in the
dialog.  Table~\ref {tab:dm} gives the number of occurrences of
discourse markers in turn initial position in the Trains corpus.  From
column two, we see that discourse markers start 4202 of the 6163
utterances in the corpus, or 68.2\%.  If we exclude turn-initial
filled pauses and acknowledgments and exclude turns that consist of
only filled pauses and discourse markers, we see that 44.1\% of the
speaker turns are marked with a non-acknowledgment discourse marker.
\begin{table}[b]
\begin{center}\small
{\small\begin{tabular}{|l|r|r|}\hline
\bf Turns that   &           &\bf Excluding initial \\
\bf start with   &\bf Number &\bf AC's and UH\_FP's \\ \hline
AC               & 3040   &  n.a. \\
CC\_D            &  824   &  1414 \\
RB\_D            &   63   &   154 \\
UH\_D            &  275   &   302 \\
UH\_FP           &  462   &  n.a. \\
Other            & 1499   &  2373 \\ \hline
Total            & 6163   &  4243 \\ \hline
\end{tabular}}
\end{center}
\caption{Discourse markers in turn-initial position}
\label{tab:dm}
\end{table}

In earlier work \cite{ByronHeeman97:eurospeech,ByronHeeman97:tr}, we
investigated the role that discourse markers play in task-oriented
human-human dialogs.  We investigated Shriffin's claim that discourse
markers can be used to express the relationship between the
information in the upcoming utterance to the information in the
discourse state \cite{Schiffrin87:book}.  For each turn that began
with a discourse marker, we coded the type of conversational move that
the discourse marker introduced.  The conversational move annotations,
described in Table~\ref {tab:cm}, attempt to capture speaker intent
rather than the surface form of the utterance.
\begin{table}
\begin{description}\setlength\itemsep{-.1cm}
\item [Restate] A restatement of either the plan or facts in the world
that have been explicitly stated before.
\item[Summarize Plan] A restatement of the current working plan where
this plan has been previously built up in pieces but has not been
previously stated in its entirety.
\item[Request for summary] Typically questions about the
total time the plan will take, such as ``what's the total on that.''
\item[Conclude] Explicit conclusion about the planning state that has
not been stated previously, e.g. `So that's not enough time' or `So we
have thirteen hours' 
\item[Elaborate Plan] Adding new plan steps onto the plan, e.g. ``How
about if we bring engine two and two boxcars from Elmira to Corning''
\item[Correction] Correcting either the plan or a misconception of the
other speaker.
\item[Respond to new info] Explicit acknowledgment of new
information, such as ``oh really'' or ``then let's do that''.   
\end{description}\vspace*{-1.5em}
\caption{Conversational move categories}\vspace*{2em}
\label{tab:cm}
\begin{center}\small
\begin{tabular}{|l|r|r|r|r|} \hline 
\multicolumn{1}{|c|} {} &
\multicolumn{4}{c|} {\bf Turns beginning with} \\ \cline{2-5}
\multicolumn{1}{|c|}{\bf Conversational Move}  & 
\multicolumn{1}{c|} {\bf And} &
\multicolumn{1}{c|} {\bf Oh} & 
\multicolumn{1}{c|} {\bf So} & 
\multicolumn{1}{c|} {\bf Well}  \\ \hline 
Restate             & 0  & 0 & 6 & 0\\ 
Summarize Plan      & 5  & 0 & 4 & 0\\ 
Request for summary & 1  & 0 & 3 & 0 \\
Conclude            & 0  & 0 & 15 & 0\\ 
Elaborate Plan      & 22 & 0 & 0 & 0 \\
Correction          & 0  & 0 & 0 & 7 \\ 
Respond to new info & 0  & 17 & 0 & 0 \\ \hline
\end{tabular}
\caption{\label{tab:information}Correlations with conversational move}
\end{center}
\end{table}
We annotated five of the Trains dialogs, containing a
total of 401 speaker turns and 24.5 minutes of speech. 

In accordance with Schiffrin, we found that utterances that summarize
information are likely to be introduced with ``so'', utterances that
add on to the speakers prior contribution (and perhaps ignore the
other conversants intervening contribution) are likely to be
introduced with ``and'', and utterances that express dissent with the
information in the discourse state are likely to be introduced with
``well''.  Table~\ref{tab:information} summarizes the co-occurrence
of turn-initial discourse markers with the conversational moves that
they introduce.  The table shows that different discourse markers
strongly correlated with particular conversational moves.  Because
discourse markers are found in turn-initial position, they can be used
as a timely indicator of the conversational move about to be made.

A more traditional method for analyzing the function of turns in a
dialog is to focus on their surface form by categorizing them into
speech acts, so we wanted to see if this sort of analysis would reveal
anything interesting about discourse marker usage in the Trains
dialogs.  Table~\ref{tab:sa_defs} defines the speech acts that were
used to annotate the dialogs.
\begin{table}
\begin{description}\setlength\itemsep{-.1cm}
\item[Acknowledge] Backchannel `Okay' or `mm-hm'.  
\item[Check] Restating old information to elicit a positive response from the 
partner (e.g. That was three hours to Bath?).
\item[Confirm] Restating old information, with no apparent intention
of partner agreement.  
\item [Filled Pause] A turn containing no information such as `hm'. 
\item [Inform] Information not previously made explicit.
\item [Request] Request for information.     
\item [Respond] Respond to a Request.   
\item [Y/N Question] Questions requiring a yes/no answer. Differ from
Check because the speaker displays no bias toward which answer he
expects. 
\item [Y/N Answer] Answering `yes', `no', `right', etc. 
\end{description}\vspace*{-1.5em}
\caption{\label{tab:sa_defs}Speech Act annotations}\vspace{2em}
\begin{center}\small
\setlength{\tabcolsep}{0.3em}
\begin{tabular}{|l|r|r|r|r|r|r|} \hline 
\multicolumn{1}{|c|}{} & 
\multicolumn{1}{c|} {\bf Total} &
\multicolumn{4}{c|} {\bf Turn begins with} & 
\multicolumn{1}{c|} {\bf DM Turns} \\ \cline{3-6}
\multicolumn{1}{|c|}{} & 
\multicolumn{1}{c|} {\bf Turns} &
\multicolumn{1}{c|} {\bf And} &
\multicolumn{1}{c|} {\bf Oh} & 
\multicolumn{1}{c|} {\bf So} & 
\multicolumn{1}{c|} {\bf Well} &
\multicolumn{1}{c|} {\bf \% of Total} \\ \hline 
\multicolumn{7}{|c|} {\bf Prior speech act initiates adjacency pair}\\ \hline
Check       & 23  & 0  & 0  & 0  & 1 & 4\%  \\ 
Request Info& 45  & 0  & 0  & 1  & 0 & 2\%  \\ 
Y/N Question& 8  & 0 & 0 & 0  & 0 & 0\% \\ \hline
\multicolumn{7}{|c|} {\bf Prior speech act concludes adjacency pair} \\ \hline
Respond     & 38  & 3  & 2  & 5  & 1& 30\%   \\
Y/N Answer  & 26 & 1  & 1  & 1  & 0& 12\%  \\ 
Acknowledge & 107 & 21 & 4  & 16 & 2 &40\%   \\ \hline
\multicolumn{7}{|c|} {\bf Prior speech act not in adjacency pair} \\ \hline
Confirm     & 42  & 2  & 0  & 0  & 1 & 7\%  \\ 
Inform      & 96  & 1  & 10 & 5  & 2 & 19\%   \\ 
Filled Pause& 6  & 0 & 0 & 0  & 0 & 0\%  \\ \hline
\end{tabular}
\caption{Prior speech act of DM-initial turns}
\label{tab:adjacency}
\end{center}
\end{table}
We found that discourse markers on the whole do not correlate strongly
with particular speech acts, as they did with conversational moves.
This is corroborated by Schiffrin's \shortcite{Schiffrin87:book}
corpus analysis, in which she concluded that turn-initiators reveal
little about the construction of the upcoming turn.  Although not
correlating with syntactic construction, discourse markers do interact
with the local discourse structure property of adjacency pairs
\cite{SchegloffSacks73:s}.  In an adjacency pair, such as
Question/Answer or Greeting/Greeting, the utterance of the first
speech act of the pair sets up an obligation for the partner to produce
the second speech act of the pair.  After the first part of an
adjacency pair has been produced, there is a very strong expectation
about how the next turn will relate to the preceding discourse,
e.g. it will provide an answer to the question just asked.

Since discourse markers help speakers signal how the current turn
relates to prior talk, we decided to investigate what speech acts
discourse markers tend to follow and how they correlate with adjacency
pairs. Table~\ref{tab:adjacency} shows the {\em prior} speech act of
turns beginning with discourse markers.  
The speech acts have been organized into those that form the first
part of an adjacency pair (Request Info, Y/N Question, and Check),
those that form second-pair-parts (Respond, Y/N/ Answer, and
Acknowledge), and those that are not part of an adjacency pair
sequence (Confirm, Inform, and Filled Pause).  The table reveals the
very low frequency of discourse marker initial turns after the
initiation of an adjacency pair.  After an adjacency pair has been
initiated, the next turn almost never begins with a discourse marker,
because the turn following the initiation of an adjacency pair is
expected to be the completion of the pair.  Since the role of that
turn is not ambiguous, it does not need to begin with a discourse
marker to mark its relationship to preceding talk.  It would indeed be
odd if after a direct question such as ``so how many hours is it from
Avon to Dansville'' the system responded ``and 6'' or ``so 6''.  A possible
exception would be to begin with ``well'' if the upcoming utterance is a
correction rather than an answer.  There is one ``so'' turn in the
annotated dialogs after a Request act, but it is a request for
clarification of the question.

After a turn that is not the initiation of an adjacency pair, such as
Acknowledge, Respond, or Inform, the next turn has a much higher
probability of beginning with a discourse marker. Also when the prior
speech act concludes an adjacency pair, the role of the next statement
is ambiguous, so a discourse marker is used to mark its relationship
to prior discourse.

In this section, we demonstrated that the choice of discourse marker
gives evidence as to the type of conversational move that the speaker
is about to make.  Furthermore, discourse markers are more likely to
be used where there are not strong expectations about the utterance that
the speaker is about to make.  Thus, discourse markers provide hearers
with timely information as to how the upcoming speech should be
interpreted.

\section{Usefulness of Discourse Markers}

We have also shown that discourse markers can be reliably identified
in task-oriented spontaneous speech.  The results given in the
previous section show that knowledge of the discourse marker leads to
strong expectations of the speech that will follow.  However, none of
the work in using machine learning techniques to predict the speech
act of the users speech has used the presence of a discourse marker.
Chu-Carroll \shortcite {Chucarroll98:aaaiss} examined syntactic type
of the utterance and turn-taking information, but not the presence of
a discourse marker.  The work of Taylor et al.~\shortcite
{Taylor-etal97:eurospeech} on using prosody to identify discourse act
type also ignores the presence of discourse markers.  Work of Stolcke
et al.~\shortcite {Stolcke-etal98:aaaiss} also ignores them.  As
Dahlb{\"a}ck and J{\"o}nsson observed \shortcite
{DahlbackJonsson92:cogsci}, it might be that speakers drop the usage
of discourse markers in talking with computer systems, but this might
be more of an effect of the current abilities of such systems and user
perceptions of them, rather than that people will not want to use
these as their perception of computer dialogue systems increases.  A
first step in this direction is to make use of these markers in
dialogue comprehension.  Machine learning algorithms of discourse acts
are ideally suited for this task.

\section{Conclusion}

In this paper, we have shown that discourse markers can be identified
very reliably in spoken dialogue by viewing the identification task as
part of the process of part-of-speech tagging and using a Markov model
approach to identify them.  The identification process can be
incorporated into speech recognition, and this leads to a small
reduction in both the word perplexity and POS tagging error rate.
Incorporating other aspects of spontaneous speech, namely speech
repair resolution and identification of intonation phrase boundary
tones, leads to further improvements in our ability to identify
discourse markers.

Our method for identifying discourse markers views this task as part
of the speech recognition problem along with POS tagging.  As such,
rather than classifying each potential word independently as to
whether it is a discourse marker or not (cf.~\cite{Litman96:jair}), we
find the best interpretation for the acoustic signal, which includes
identifying the discourse markers.  Using this approach means that the
probability distributions that need to be estimated are more
complicated than those traditionally used in speech recognition
language modeling.  Hence, we make use of a decision tree algorithm to
partition the training data into equivalence classes from which the
probability distributions can be computed.

Automatically identifying discourse markers early in the processing
stream means that we can take advantage of their presence to help
predict the following speech.  In fact, we have shown that discourse
markers not only can be used to help predict how the speaker's
subsequent speech will build on to the discourse state, but also are
often used when there are not already strong expectations, in terms of
adjacency pairs.  However, most current spoken dialogue systems ignore
their presence, even though they can be easily incorporated into
existing machine learning algorithms that predict discourse act types.

\section{Acknowledgments}

This material is based upon research work supported by the NSF under
grant IRI-9623665 and by ONR under grant N00014-95-1-1088 at the
University of Rochester.

\bibliographystyle{aaai}


\def\thebibliography#1{\section*{References}\list
 {}{\setlength{\labelwidth}{0pt}\setlength{\leftmargin}{\parindent}
 \setlength{\itemindent}{-\parindent}}
 \itemsep=-2pt
 \def\baselinestretch{0.95}\small
 \vskip -9pt
 \advance\leftmargin\labelsep
 \def\newblock{\hskip .11em plus .33em minus -.07em}
 \sloppy\clubpenalty4000\widowpenalty4000
 \sfcode`\.=1000\relax}

\end{document}